\documentclass[10pt,journal,compsoc]{IEEEtran}
\usepackage{times}
\usepackage{epsfig}
\usepackage{graphicx}
\usepackage{amsmath}
\usepackage{amssymb}
\usepackage{color}
\usepackage{enumitem}
\usepackage{pifont}
\usepackage[hyphens]{url}
\usepackage{hyperref}
\usepackage{floatrow}
\usepackage{cite}

\newcommand{\cmark}{\ding{51}}%
\newcommand{\xmark}{\ding{55}}%
\newcommand{\comment}[1]{}
\newcommand{\boldstartspace}[1]{\vspace{0.1in}\noindent\textbf{#1}}
\newcommand{\edit}[1]{#1}
\newcommand{\KL}[1]{}
\newcommand{\RR}[1]{}
\newcommand{\LX}[1]{}

\newcommand{\arxiv}[1]{} 

\usepackage{xcolor}
\hypersetup{
    colorlinks,
    linkcolor={red!90!black},
    citecolor={blue!90!black},
    urlcolor={blue!90!black},
    breaklinks=true
}

\begin{document}
%
\title{View Synthesis of Dynamic Scenes based on Deep 3D Mask Volume}
%
%
%
%

\author{Kai-En Lin*, Guowei Yang*,\thanks{* denotes equal contribution.} Lei Xiao, Feng Liu, Ravi Ramamoorthi\arxiv{,~\IEEEmembership{Fellow,~IEEE}}
 
  \thanks{
   \textbullet \quad K. Lin, G. Yang and R. Ramamoorthi are with the CSE Department at the University of California, San Diego, at La Jolla, CA 92037. Email: \{k2lin@, g4yang@, ravir@cs \}.ucsd.edu } 
   
   \thanks{ \textbullet \quad Lei Xiao is with Reality Labs Research at Meta, at Redmond, WA 98052 Email: lei.xiao@fb.com}
   
  \thanks{ \textbullet \quad Feng Liu is with the Computer Science Department at Portland State University, at Portland, OR 97207 Email: fliu@cs.pdx.edu}
 
}

\IEEEtitleabstractindextext{%
\begin{abstract}
Image view synthesis has seen great success in reconstructing photorealistic visuals, thanks to deep learning and various novel representations.
The next key step in immersive virtual experiences is view synthesis of dynamic scenes.
However, several challenges exist due to the lack of high-quality training datasets, and the additional time dimension for videos of dynamic scenes.
To address this issue, we introduce a multi-view video dataset, captured with a custom 10-camera rig in 120FPS.  The dataset contains 96 high-quality scenes showing various visual effects and human interactions in outdoor scenes.  We develop a new algorithm, Deep 3D Mask Volume, which enables temporally-stable view extrapolation from binocular videos of dynamic scenes, captured by static cameras.
Our algorithm addresses the temporal inconsistency of disocclusions by identifying the error-prone areas with a 3D mask volume, and replaces them with static background observed throughout the video.  Our method enables manipulation in 3D space as opposed to simple 2D masks, 
We demonstrate better temporal stability than frame-by-frame static 
view synthesis methods, or those that use 2D masks.
The resulting view synthesis videos show minimal flickering artifacts and allow for larger translational movements.


\comment{
Image view synthesis has seen great success in reconstructing photorealistic visuals, thanks to deep learning and various novel representations.
On the other hand, video view synthesis remains relatively unexplored due to the lack of high-quality training datasets, and the additional time dimension for dynamic scenes.
One of the main challenges is to handle disocclusion in a temporally-stable way.
To solve this, we introduce a multi-view video dataset, captured with a custom camera rig, and a novel 3D mask volume to identify and replace the error-prone disoccluded areas with background observations from other temporal frames.
We create temporal plane sweep volumes from binocular input image sequences, and use them to produce a 3D mask volume that is able to reason about the disocclusion and depth differences to replace the erroneous renderings of dynamically-occluded regions with stable and accurate renderings from static background multiplane images. 
Our method achieves better temporal stability compared to frame-by-frame static view synthesis methods on our new multi-view video dataset, which contains various dynamic human interactions, visual effects and different outdoor scenes.
}

\end{abstract}


\begin{IEEEkeywords}
Computer Vision, View Synthesis
\end{IEEEkeywords}}

\maketitle

\IEEEdisplaynontitleabstractindextext

%
\IEEEpeerreviewmaketitle


%
%
%
%

 

\IEEEraisesectionheading{\section{Introduction}\label{sec:introduction}}
\begin{figure*}
\centering
\includegraphics[width=1.0\textwidth]{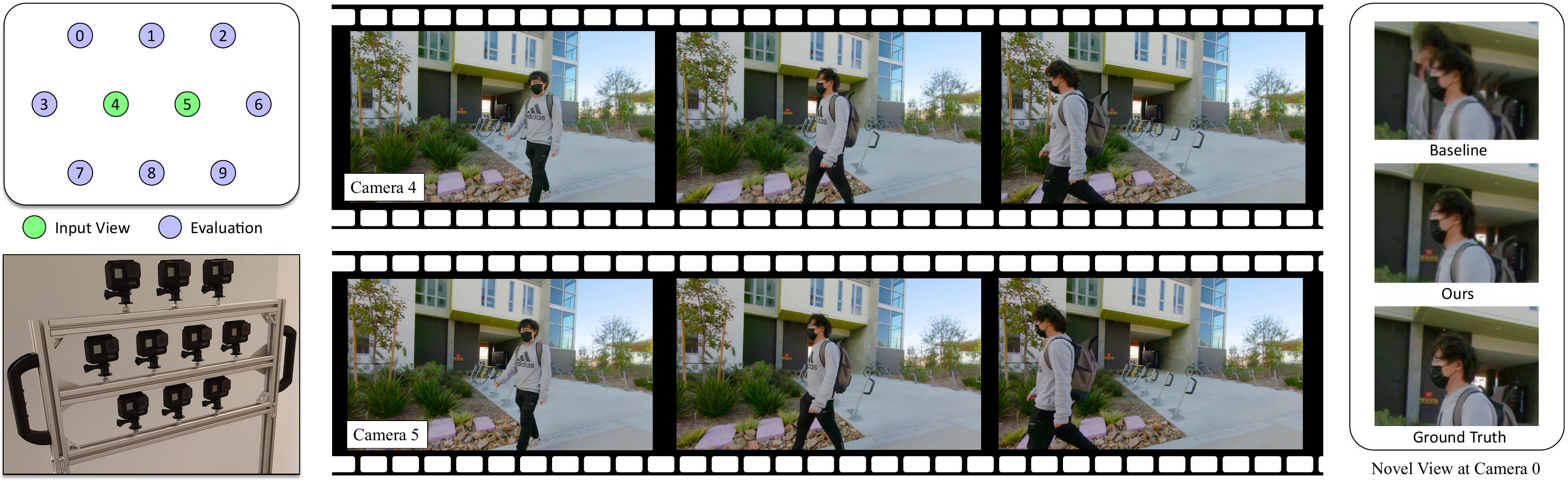}
\caption{Our custom camera rig. Top left figure shows the configuration we use for evaluation in Sec. \ref{sec:experiments}.
We show the input stereo image sequences from camera 4 and camera 5 in the middle.
The rightmost column shows the crops of rendered novel view at camera 0.
Artifacts appear when the novel view is translated by a larger distance.
We use the conventional MPI method \cite{mildenhall2019llff} as our baseline algorithm.
Note how the area on top of the person's head is distorted and shows ``stack of cards'' artifacts.
This type of artifact flickers in a dynamic video as the network hallucinates the disocclusion per-frame.}
\label{fig:camera_rig}
\end{figure*}



\IEEEPARstart{R}{ecent} advances in view synthesis have shown promising results in creating immersive virtual experiences from images.
Nonetheless, in order to reconstruct compelling and intimate interaction with the virtual scene, the ability to incorporate temporal information is much needed.
In this paper, we study a specific setup where the input videos are from static, binocular cameras and novel views are mostly extrapolated from the input videos, similar to the case in StereoMag\cite{zhou2018stereo}.
We believe that this case is useful as dual- and multi-camera smartphones are gaining traction and it could also prove to be interesting for 3D teleconferencing, surveillance or playback on virtual reality headsets.
Moreover, we can acquire the dataset from a static camera rig as shown in Fig.\ref{fig:camera_rig}.
Although we can apply state-of-the-art image view synthesis algorithms \cite{zhou2018stereo, SrinivasanCVPR2019, mildenhall2019llff, Shih3DP20} on each individual video frame, the results lack temporal consistency and often show flickering artifacts.
The issues mostly come from the unseen occluded regions as the algorithm predicts them on a per-frame basis. 
The resulting estimations are not consistent across the time dimension, which causes some regions to become unstable when shown in a video.



In this paper, we address the temporal inconsistency when extrapolating views by exploiting the static background information across time.
To this end, we employ a 3D mask volume, which allows manipulation in 3D space as opposed to a 2D mask, to reason about moving objects in the scene and reuse static background observations across the video.
As shown in Fig.\ref{fig:algorithm}, we first promote the instantaneous and background inputs into two sets of multiplane images (MPI)\cite{zhou2018stereo} via an MPI network.
Then, we warp the same set of input images to create a temporal plane sweep volume, providing information about the 3D structure of the scene.
The mask network converts this volume to a 3D mask volume which allows us to blend between the two sets of MPIs. 
Finally, the blended MPI volume can render novel views with minimal flickering artifacts.


To train this network, we also introduce a new multi-view video dataset to address the lack of publicly available data. We build a custom camera rig comprised of 10 action cameras and capture high-quality 120FPS videos with the static rig (see Fig. \ref{fig:camera_rig}).
Our dataset contains 96 dynamic scenes of various outdoor environments and human motions.
We show that the proposed method generates temporally stable results against previous state-of-the-art methods, while only using two input views.


Our contributions can be summarized as:
\begin{itemize}[noitemsep,topsep=0pt]
    \item a multi-view video dataset composed of 96 dynamic scenes (Sec. \ref{sec:dataset});
    \item a novel 3D volumetric mask able to segment dynamic objects from static background in 3D, producing higher-quality and temporally stable results than state-of-the-art methods (Sec. \ref{sec:3d_mask_volume});
    \item a synthetic dataset to evaluate complex background (Sec.~\ref{sec:synthetic});
    \item experiments including recent NeRF-based dynamic view synthesis methods (Sec.~\ref{sec:comparison}, Sec.~\ref{sec:synthetic}).
\end{itemize}

\edit{This paper is an extended version of Deep 3D Mask Volume for View Synthesis of Dynamic Scenes~\cite{lin2021deep}.
In this \arxiv{journal }version, we conduct further experiments to evaluate concurrent NeRF-based methods~\cite{du2021nerflow,li2020neural,tretschk2021nonrigid} in Sec.~\ref{sec:comparison}.
These methods target a monocular dynamic camera setup different from our static stereo camera setup.
A moving monocular camera effectively provides multiple viewpoints of the static scene components.
On the contrary, static stereo cameras can only supply two viewpoints and thus their methods do not perform as well as our proposed method.
To show experiments in a more controlled environment and allow for more complex backgrounds, we created a new synthetic dataset to evaluate the performance in Sec.~\ref{sec:synthetic}.
We demonstrate how our method can tackle the dynamic background with multiple actors.
Moreover, we detail how different loss functions would affect the visual results in Sec.~\ref{sec:ablation}, as well as large distance view extrapolation in Sec.~\ref{sec:large_dist} and extension to more input views in Sec.~\ref{sec:more_views}.}
\section{Related Work}
Our goal is to achieve temporally stable view synthesis on dynamic scenes. 
We are inspired by several previous methods in view synthesis and space-time synthesis.

\subsection{View synthesis}
View synthesis is a complicated problem which has become a popular field of research in computer vision and graphics.
Earlier lines of work utilize dense sampling from the scene to create light fields \cite{gortler1996lumigraph,levoy1996light}.
Image-based rendering techniques \cite{buehler2001unstructured,debevec1996modeling} exploit proxy geometry of the scene to produce novel view renderings.
Later extensions on this topic introduce better modeling of the scene structure \cite{shade1998layered} and hand-crafted heuristics \cite{davis2012unstructured,penner2017soft}.
As deep learning became dominant, learning-based methods \cite{hedman2018deep,flynn2016deepstereo,kalantari2016learning,10.1145/3306346.3323007,niklaus20193d} have shown promising results.
Recently, a class of research works focuses on combining novel representations \cite{zhou2018stereo,mildenhall2020nerf,mildenhall2019llff,SrinivasanCVPR2019,flynn2019deepview,jiang2020local,lombardi2019neural,sitzmann2019deepvoxels,park2019deepsdf,lin2020mdp} with a differentiable rendering pipeline to produce high-quality results.
Another exciting advance is neural radiance fields (NeRF) \cite{mildenhall2020nerf}, which encodes the 3D scene structure in a compact continuous 5D volumetric function.
Although NeRF has shown promising view synthesis results, it has to overfit to the given scene with enough samples (10 or more), requiring time-consuming per-scene training.
Rendering time could take up to 30s for one image, whereas our pipeline allows inference and rendering in less than 2s without dedicated optimization, using only binocular input views.

Instead, in this paper we focus on a specific layered representation, multiplane images (MPI) \cite{szeliski1998stereo, zhou2018stereo, SrinivasanCVPR2019, mildenhall2019llff, broxton2020immersive}, as it provides good generalizability across various scenes and efficiency capable of real-time rendering.
Our proposed method directly tackles the temporal instability introduced in MPIs when the disoccluded areas lead to different estimations across time. 

\subsection{Space-time synthesis}
Space-time synthesis is a more complicated problem since it not only involves movement of the novel viewpoint in space, but also incorporates differences of time.
A body of work covers appearance changes such as relighting while changing views \cite{10.1145/3197517.3201313, 10.1145/3306346.3323007, bi2020deep, bi2020neural, meshry2019neural}.
However, these methods focus on the lighting change with respect to a static scene, treating dynamic objects in the scene as outliers.
On the other hand, some methods directly target dynamic scenes \cite{broxton2020immersive,bansal20204d,Bemana2020xfields,yoon2020novel,zitnick2004high}.
While our method utilizes MPIs similar to Broxton et al.\cite{broxton2020immersive}, they employ dense sampling of 46 cameras to reconstruct light fields of the viewing volume, essentially interpolating between cameras.
Our method focuses on the stereo case similar to StereoMag\cite{zhou2018stereo}, targeting extrapolation from stereo inputs like dual-camera smartphones.
In addition, unlike depth-based methods \cite{bansal20204d,yoon2020novel}, we do not require any explicit depth maps to render novel viewpoints.
As depth-based methods often yield flickering and require hole-filling, we instead use a representation more suitable for rendering.
Another issue that these methods do not address is the lack of generalizability.
Bansal et al. \cite{bansal20204d} is trained on limited data which could make the learned network overfit to a small number of scenes.
Moreover, while Yoon et al. \cite{yoon2020novel} uses a pretrained network to ensure generalizability on unseen scenes, it still requires human-generated masks for foreground and background separation.
We capture various dynamic scenes with human interactions to train our network and ensure that it is generalizable across different unseen scenes.
Also, our network utilizes the background information extracted from video and uses it to directly segment the foreground and background in 3D space without any human input.

\edit{Concurrently, there are several NeRF-based algorithms \cite{du2021nerflow,li2020neural,tretschk2021nonrigid} which demonstrate state-of-the-art performance on monocular video inputs with a moving camera.
For static parts of the scene, a moving camera provides multi-view cues to the network and they can be reconstructed in the same process as the original NeRF~\cite{mildenhall2020nerf}.
For dynamic parts of the scene, NSFF~\cite{li2020neural} learns an implicit representation of the scene flow and warps the sampled points to render the scene at different timesteps.
Similarly, NeRFlow~\cite{du2021nerflow} also uses an MLP network to learn the underlying scene flow but it incorporates a neural ODE to enforce consistency across continuous time. Non-rigid NeRF~\cite{tretschk2021nonrigid} optimizes for a canonical volume model, then it uses deformation fields to generate renderings at different timestamps.
Although these methods work well for a single moving camera, they are not able to acquire good 3D geometry for a pair of static cameras.
As demonstrated in Sec.~\ref{sec:comparison}, our MPI-based method is able to utilize better geometry priors to provide high-quality results during extrapolation with less distortion and flickering.
}

\begin{table*}[t]
  \centering
  \resizebox{1.0\textwidth}{!}{%
  \begin{tabular}{l r r r r r r r}
    Dataset & Scene count & Rigid rig & Large disparity & Views & Dynamic & Public & Remarks \\
    \hline
    Real Forward-Facing \cite{mildenhall2019llff} & 65 & \xmark & \cmark & 25 & \xmark & \xmark & Loosely gridlike formation \\
    Spaces \cite{flynn2019deepview} & 100 & \cmark & \cmark & 16 & \xmark & \cmark & Strictly gridlike formation\\
    Immersive LF Video \cite{broxton2020immersive} & 130 & \cmark & \cmark & 46 & \cmark & \xmark & Spherical formation\\
    Dynamic Scene \cite{yoon2020novel}  & 8 & \cmark & \cmark & 12 & \cmark & \cmark & Few temporal frames \\
    Single Image LF \cite{Li2020LF} & $\sim$2000 & \cmark & \xmark & 196 & \xmark & \cmark & Small baseline light fields \\
    RealEstate10K \cite{zhou2018stereo} & $\sim$10000 & \xmark & \cmark & 1 & \xmark & \cmark & Static scenes \\
    Open4D \cite{bansal20204d} & 6 & \xmark & \cmark & 15 & \cmark & \cmark & Free-viewpoint capture \\
    MannequinChallenge \cite{li2019learning} & $\sim$2000 & \xmark & \cmark & 1 & \xmark & \cmark & Mostly static scenes \\
    X-Fields \cite{Bemana2020xfields} & 8 & \cmark & \cmark & 5 & \cmark & \cmark & Few temporal frames \\
    KITTI \cite{Menze2015CVPR} & 400 & \cmark & \cmark & 2 & \cmark & \cmark & Binocular setup on cars \\
    \hline
    \textbf{Ours} & 96 & \cmark & \cmark & 10 & \cmark & \cmark & Publicly released \\
  \end{tabular}
  }
  \caption{Comparison of different multi-view datasets.}
  \label{tab:dataset}
\end{table*}
\section{Dataset}\label{sec:dataset}

High-quality video datasets are crucial for learning-based novel-view video synthesis algorithms. The ideal datasets would contain a diversity of scenes, captured at multiple synchronized views. 
In this work we introduce a novel multi-view video dataset. We discuss the limitations of existing datasets compared to our dataset in Sec. \ref{sec:multi-view}.
We describe our data capture and generation process in Sec. \ref{sec:generation}.
Finally, we discuss the statistics and advanced properties of our dataset in Sec. \ref{sec:statistics}.

\subsection{Multi-view video dataset}\label{sec:multi-view}

\begin{figure}
    \floatbox[{\capbeside\thisfloatsetup{capbesideposition={right,top},capbesidewidth=4cm}}]{figure}[0.4\textwidth]
    {\caption{Digital clock and the randomly moving QR code pattern used to perform synchronization. We have two ways to do synchronization: (1) matching the timestamp; (2) aligning the QR code location in all views.
    We use these methods to ensure the synchronization is accurate enough.}
    \label{fig:qrcode}}
    {\includegraphics[width=0.32\textwidth]{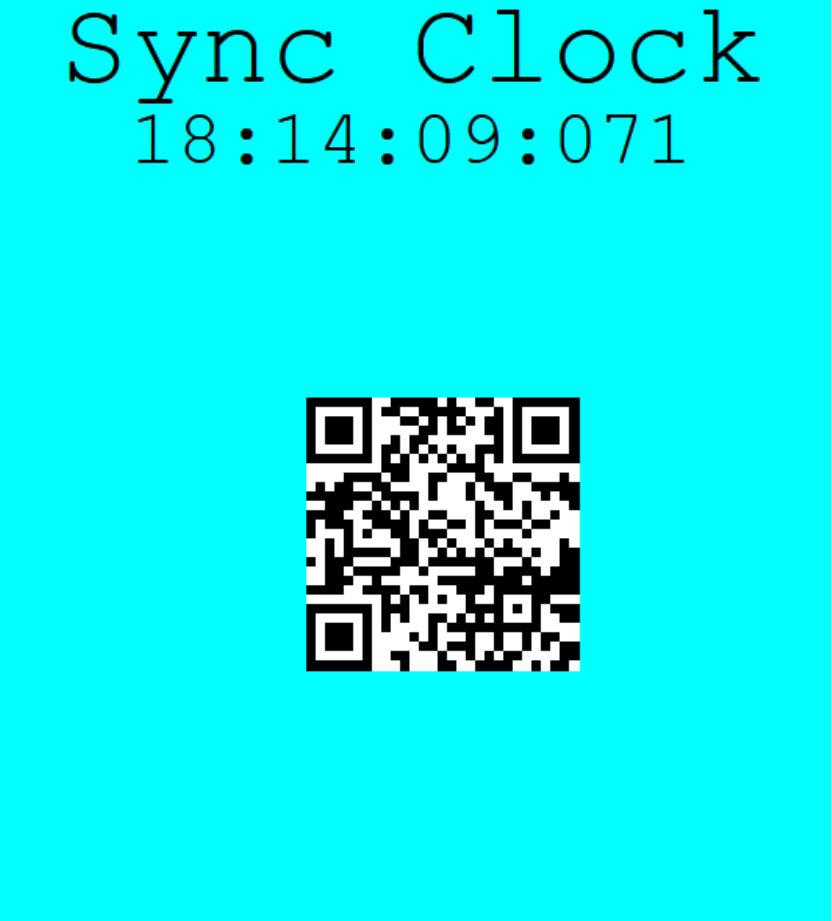}}
\end{figure}

As shown in Table \ref{tab:dataset}, we evaluate several properties which are important to train a generalized view synthesis network.
Specifically, a rigid camera rig is preferred as it can provide good pose priors and ensure the accuracy of the estimated camera poses.
On the contrary, unstructured captures like Real Forward-Facing \cite{mildenhall2019llff} and Open4D \cite{bansal20204d} do not use pose priors and utilize structure from motion, which could produce varying accuracy depending on scene geometry and the texture presented.
In addition, rigid camera rigs allow for capture of dynamic scenes with multiple simultaneous camera views.
On account of the above reasons, our dataset is captured with a custom camera rig that is rigid and robust enough to offer good pose priors.

Number of views is also an important factor for a multi-view dataset since different combinations of input and target camera pairs provide diversity in baselines and camera motions.
X-Fields \cite{Bemana2020xfields} and KITTI \cite{Menze2015CVPR} provide limited views and camera motions and thus are not as useful for video view synthesis tasks.
Our dataset offers 10 different camera views in a gridlike formation (see Fig. \ref{fig:camera_rig}). For our binocular view synthesis task, we choose 2 views out of 10 and 1 from the rest to construct a training pair.

The most important feature is to have enough temporal frames and dynamic movements for training.
Most datasets fail at this part as they target the image view synthesis task instead of a video one.
Although the Dynamic Scene Dataset presented by Yoon et al.\cite{yoon2020novel} targets the dynamic scenes, it uses frame skips to keep salient movements.
Thus, the movements shown in the dataset are not smooth and fail to provide enough training samples.
To address this issue, our dataset is captured in 120 FPS and synchronized as a post-process (see Sec. \ref{sec:generation}), making it easy to perform and evaluate view synthesis at different framerates.

One dataset that targets the purpose of video view synthesis is the Immersive Light Field Video dataset proposed by Broxton et al.\cite{broxton2020immersive}, which contains 46 camera views and 130 different dynamic scenes.
However, the full dataset is not publicly available to the community.
Our full dataset can be found at \url{http://cseweb.ucsd.edu/\%7eviscomp/projects/ICCV21Deep/}

\subsection{Dataset generation}\label{sec:generation}
Our video dataset is captured with a custom camera rig that consists of 10 GoPro Hero 7 Black action cameras as shown in Fig. \ref{fig:camera_rig}.
The horizontal baseline between neighbor cameras is approximately 10 cm and the vertical distance between rows is around 14 cm.
We captured 96 outdoor videos in 120 FPS, with the camera rig being static for each video.
As GoPros only allow fisheye mode for high FPS captures, we calibrate the cameras with a 17x14 checkerboard pattern (squares have side lengths of 40mm) and undistort the videos using a pinhole camera model \cite{10.5555/580035} implemented in OpenCV \cite{opencv_library}.
For camera extrinsics, we choose the first frame from all views as inputs to COLMAP \cite{schoenberger2016mvs, schoenberger2016sfm}, which then does feature extraction, feature matching, and sparse reconstruction.
The reconstructed camera poses are assumed to be fixed for the duration of each video. 
In addition, to achieve synchronization, we display a digital clock with randomly appearing QR code patterns (see Fig. \ref{fig:qrcode}) on a high refresh rate screen that can be seen by all cameras at the same time.
Then, we manually edit and align the multi-view videos according to the digital clock and QR code pattern.


\begin{table}[t]
  \caption{Number of videos that contain each occlusion type as described in Sec.~\ref{sec:statistics}. Note that most scenes typically contain multiple types of occlusion.}
  \label{tab:statistics}
  {\begin{tabular}{c | c c c c | c}

    \textbf{Occlusion Types} & \textbf{(a)} & \textbf{(b)} & \textbf{(c)} & \textbf{(d)} & \textbf{Total videos} \\
    \hline
    Count & 90 & 96 & 42 & 19 & 96  \\
    \hline
  \end{tabular}}
\end{table}

\begin{figure*}
\centering
\includegraphics[width=0.95\textwidth]{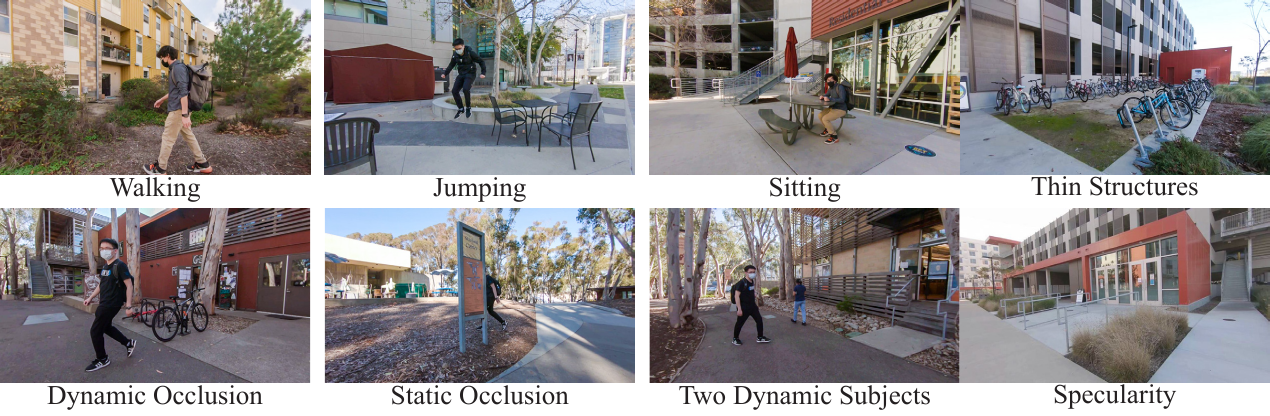}
\caption{A selection of still frames from our dataset. We captured various dynamic scenes with human motions, including walking, running, jumping and sitting down. Note that cameras remain static for the whole duration of the capture. {\LX{Make the subfigure captions font larger.}}}
\label{fig:scene}
\end{figure*}

\begin{figure*}
\begin{center}
\includegraphics[width=\linewidth]{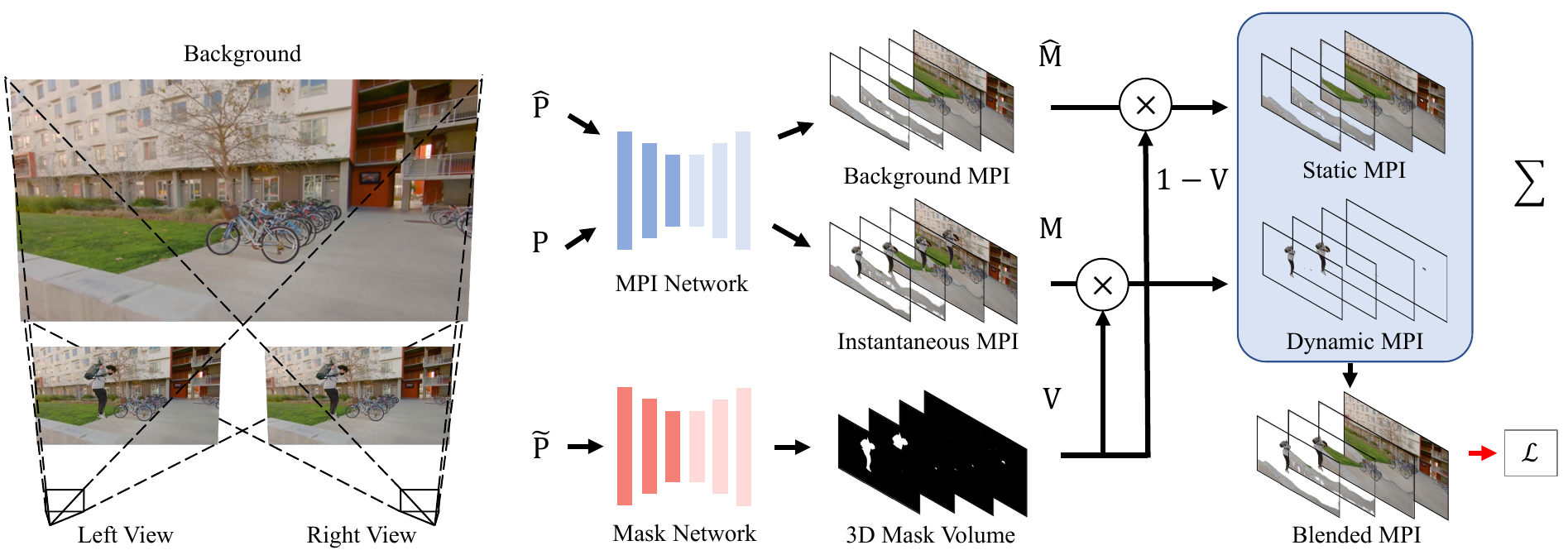}
\end{center}
   \caption{Overview of our pipeline. Given binocular input videos, our MPI network promotes the 2D multiview images to two 3D MPI representations; one encodes the instantaneous information and the other encodes the background information. The mask network produces a 3D mask volume $\textbf{V}$ to modulate the MPIs and blend them together, producing the final output. Please see Sec. \ref{sec:3d_mask_volume} for more details.}
\label{fig:algorithm}
\end{figure*}

\subsection{Dataset statistics}\label{sec:statistics}




Our videos are mostly around 1 to 2 minutes long and all videos are shot in 120 FPS.
We cover different scenes to ensure that the surface reflectance variety is high enough.
For example, in Fig.\ref{fig:scene} we show that in our dataset we cover different buildings, furniture, foliage and specularity effects.
Another important aspect of our dataset is the inclusion of different human motions, including slower motions like walking, sitting down and faster motions, such as running, jumping and arms waving.
We now discuss four possible types of occlusion interactions and show the numbers of their occurrences in Table~\ref{tab:statistics}.

\boldstartspace{(a) Static occluder and static background.}
Most view synthesis methods target this case as this is one of the most common cases. 
We desribe it as a static occluder in the scene blocking the line-of-sight from the cameras to the background scene.
For instance, the table in the sitting scene shown in Fig. \ref{fig:scene} occludes the areas behind. 
Background information can only be acquired from the views with direct line-of-sight.
As such, it is difficult to recover the unseen regions without prior knowledge of the scene.
However, temporal consistency in these areas is easily achievable because inputs remain relatively unchanged throughout the video.
Hallucination of the disoccluded areas can also remain the same for this case.

\boldstartspace{(b) Dynamic occluder and static background.}
Another type of event happens when a dynamic object is moving across the scene.
For example, when a person is walking through the scene, the camera has line-of-sight on the background behind the person at some point in the video.
In this case, it is relatively easy to acquire static background information as the occluder does not block the line-of-sight in all video frames.
Combining information from multiple frames throughout the video provides an accurate rendering of what is behind the dynamic occluder.
Temporal consistency in this case can also be maintained by substituting the static background for the dynamically-occluded regions.
In other words, we can perform hole-filling based on the observations from other video frames.
Our proposed method takes advantage of this prior knowledge to generate temporally-stable view synthesis results, as opposed to previous methods.

\boldstartspace{(c) Static occluder and dynamic background.}
This case happens when an object moves behind a static occluder and thus the camera does not have full visuals on it.
For instance, a person walks behind a traffic sign or a wall.
In the traffic sign case, as it is only a short-term occlusion, the person's appearance can be interpolated between different frames.
However, in the case of a larger wall, this becomes difficult to solve as extrapolating the movement is complicated and the ambiguity could lead to different outcomes.
In general, it is difficult to accurately predict the trajectory of the occluded object without assuming it is moving at constant velocity.
For temporal consistency, the movement of dynamic objects can lead to instability of the novel view prediction.
Our method learns to detect the dynamic movements and treat the static part of the scene as (a) such that flickering artifacts are kept at a minimal level.


\boldstartspace{(d) Dynamic occluder and dynamic background.}
The last case happens when the occluder and the background object are both moving or the background appearance is changing.
For instance, this can happen when two people are walking in the opposite direction parallel to the camera's image plane.
Similar to (c), how the occluded object is moving remains ambiguous and hard to resolve deterministically.
Although we do not have a clear idea of the occluded parts, we can still ensure it is temporally stable when shown.
We can reduce this case to (b) with the ambiguity that the occluded object can move anywhere.
And as a result, the occluded regions look more or less similar to the static background.

Our dataset contains diverse occlusion interactions and we show results in Sec.\ref{sec:comparison} and provide an analysis in Fig.\ref{fig:analysis}.

\section{Deep 3D Mask Volume}



Our goal is to synthesize temporally consistent novel view videos given stereo video inputs.
Consequently, we build our algorithm upon prior work on multiplane images \cite{zhou2018stereo, mildenhall2019llff} and propose a novel mask volume structure to fully utilize the temporal background information and the layered representation.
In this section, we start with a brief review of the multiplane images in Sec. \ref{sec:mpi}.
Then we describe our 3D mask volume in Sec. \ref{sec:3d_mask_volume}.
Finally we discuss our loss function design in Sec. \ref{sec:loss_function}.
Please refer to Fig.\ref{fig:algorithm} for an overview of our algorithm pipeline.

\subsection{Multiplane images}\label{sec:mpi}
Our approach takes inspiration from recent advancements in multiplane image representation \cite{szeliski1999stereo,zhou2018stereo}.
Multiplane images (MPI) are a layered representation of the 3D scene.
They consist of $D$ layers of RGB$\alpha$ images, representing the viewing frustum from the perspective of a virtual reference camera.
The planes partition the viewing frustum according to equally-spaced disparity (inverse depth) values $d_0, d_1, ..., d_{D-1}$. 
Each layer of the MPI encodes color $C$ and transparency information $\alpha$ at a specified plane depth $d$.
We denote the MPI layer at disparity $d$ as a tuple of $(C_d, \alpha_d)$.
To construct such a volume, we warp input views to the reference camera position to construct a plane sweep volume (PSV).
The PSV is then used as the input to a 3D CNN similar to the one used by Mildenhall et al.\cite{mildenhall2019llff} and it generates the corresponding MPI volume.
To render a novel viewpoint $j$ from camera $i$, the MPI layers are warped using planar homography as follows:
\begin{equation}
    \mathcal{W}_{i\rightarrow j}^d(C_d, \alpha_d),    
\end{equation}
where $\mathcal{W}$ is the warping operator.
The warped MPIs are then composited with the over operation.
To be more specific, we calculate the per-pixel transmittance $t$ from the alpha value at location $(x, y)$ on plane $d$ by
\begin{equation}
    t(x, y, d) = \alpha(x, y, d)\prod_{d'>d}[1-\alpha(x, y, d')].
\end{equation}
The final rendering at each pixel $C_{final}$ is computed as
\begin{equation}
    C_{final}(x, y) = \sum_d{C(x, y, d)\alpha(x, y, d)\prod_{d'>d}[1-\alpha(x, y, d')]}.
\end{equation}
These computations are parallelizable and their efficiency during rendering makes the MPI a good representation for fast view synthesis.

One observation of MPIs is that the unseen parts in the volume are often merely repeated texture of the foreground objects~\cite{SrinivasanCVPR2019}.
This happens when the input camera baseline is not large enough and the resulting PSV cannot provide further information about the background. In addition, these areas typically present different estimations between frames. Therefore, the unseen areas produce visible artifacts, especially in video view synthesis (see Fig.~\ref{fig:camera_rig}). On the other hand, visible parts usually provide temporally stable results as can be seen in Broxton et al. \cite{broxton2020immersive}


\subsection{3D mask volume generation}\label{sec:3d_mask_volume}
From Sec.~\ref{sec:mpi}, we observe that most artifacts are introduced by the disocclusion of moving objects.
In order to address this issue, we seek to find a 3D mask volume that identifies the dynamic components and removes the flickering artifacts behind them accordingly.
To be more specific, given a pair of stereo image sequences of length $n$, $\{\textbf{I}^L_{0}, \textbf{I}^L_{1}, ..., \textbf{I}^L_{n-1}\}$ and $\{\textbf{I}^R_{0}, \textbf{I}^R_{1}, ..., \textbf{I}^R_{n-1}\}$, we wish to derive a 3D mask $\textbf{V}(x, y, d)$, such that
\begin{equation}\label{eq:mask_volume}
    \textbf{V}(x, y, d) = \begin{cases}
                            1, & \text{if } \textbf{I}(x, y) \neq \hat{\textbf{I}}(x, y), d > \textbf{D}(x, y) \\
                            0, & \text{otherwise}
                          \end{cases},
\end{equation}
where $\textbf{I}$ is the instantaneous frame, $\hat{\textbf{I}}$ denotes the background image, and $\textbf{D}$ is the scene disparity observed by the camera.
We drop the frame subscript as a shorthand for instantaneous frame in the following discussion.
In addition, we represent the instantaneous MPI of the scene as $\textbf{M}(x, y, d)$, and the background MPI as $\hat{\textbf{M}}(x, y, d)$.

The main purpose of the 3D mask volume $\textbf{V}(x, y, d)$ is to partition the scene $\textbf{M}(x, y, d)$ into two parts: static and dynamic.
The static portion of the MPI does not change for the whole video duration, and thus $\textbf{M}(x, y, d) = \hat{\textbf{M}}(x, y, d), \text{when } \textbf{V}(x, y, d) = 0$. 
The synthesized novel view of these parts is temporally stable and requires no further modification to the algorithm.
On the contrary, the dynamic objects ($\textbf{V}(x, y, d) = 1$) could move in different directions.  The disoccluded areas, given mathematically by $\textbf{M}(x, y, d) \text{ if } \textbf{I}(x, y) \neq \hat{\textbf{I}}(x, y), d > \textbf{D}(x, y)$, often change with them, producing ``stack of card" artifacts and flickering when viewed from another angle (see Fig. \ref{fig:camera_rig}).
However these areas in fact usually resemble the background $\hat{\textbf{I}}$.
With this knowledge, a clear separation between the static and dynamic scene components allows us to identify the disocclusion and minimize the temporal inconsistency by
\begin{equation}\label{eq:remove_disocc}
    \textbf{M}(x, y, d) \xleftarrow{} \hat{\textbf{M}}(x, y, d) \\ \text{ if } \textbf{I}(x, y) \neq \hat{\textbf{I}}(x, y), d < \textbf{D}(x, y).
\end{equation}
Essentially, we are using the temporally-stable static background to replace the unknown disoccluded areas.
An illustration of the mask is given in Fig. \ref{fig:algorithm}.

In order to perform the operation in Eq. \ref{eq:remove_disocc}, our network is composed of two networks:
\textbf{MPI network} generates 2 layered representations of the 3D scene, namely $\textbf{M}(x, y, d)$ and $\hat{\textbf{{M}}}(x, y, d)$;
\textbf{Mask network} produces the 3D mask volume $\textbf{V}(x, y, d)$ satisfying Eq. \ref{eq:mask_volume}. 
We show each network in Fig.\ref{fig:algorithm} and discuss them in details as follows:

\boldstartspace{MPI network.} It is necessary to acquire 3D information from both the instantaneous frame and throughout the whole video, so we can then obtain the needed information behind the dynamic occluder.
To this end, we first apply a median filter $\mathcal{A}$ on the image sequences
\begin{equation}\label{eq:median_filter}
    \hat{\textbf{I}} = \mathcal{A}(\{\textbf{I}_0, \textbf{I}_1, ..., \textbf{I}_{n-1}\}).
\end{equation}
It is applied to both views to generate the corresponding background images.


Then, we can inversely warp $\textbf{I}^R$ and $\hat{\textbf{I}}^R$ to the left camera and construct a PSV.
The PSV from the instantaneous frame is generated as
\begin{equation}
    \textbf{P} = \{\mathbf{I}^L, \mathcal{W}^{d_0}_{R\rightarrow L}(\textbf{I}^R), \mathcal{W}^{d_1}_{R\rightarrow L}(\textbf{I}^R), ..., \mathcal{W}^{d_{D-1}}_{R\rightarrow L}(\textbf{I}^R)\}.
\end{equation}
It is then used as an input to a 3D CNN $\mathcal{F}_{\theta}$ to produce the instantaneous MPI, $\textbf{M} = \mathcal{F}_{\theta}(\textbf{P})$.
Similarly, we construct the background MPI, $\hat{\textbf{M}}$, using another PSV, $\hat{\textbf{P}}$, generated from $\hat{\textbf{I}}^L$ and $\hat{\textbf{I}}^R$.
The two MPIs, $\textbf{M}$ and $\hat{\textbf{M}}$, now contain the information of the dynamic occluder and the static background.

\boldstartspace{Mask network.} We utilize another 3D CNN $\mathcal{G}_\theta$ to reason about the relationship between the MPIs and generate a mask volume $\textbf{V}$ to satisfy Eq. \ref{eq:mask_volume}.
Inspired by background matting \cite{BGMv2} on 2D images, our mask network takes a similar approach but in 3D space.
From Eq. \ref{eq:median_filter}, we define a temporal plane sweep volume (TPSV) as follows
\begin{multline}
    \Tilde{\textbf{P}} = \{\mathbf{I}^L, \mathcal{W}^{d_0}_{R\rightarrow L}(\textbf{I}^R), ..., \mathcal{W}^{d_{D-1}}_{R\rightarrow L}(\textbf{I}^R), \\ 
                       \hat{\mathbf{I}}^L, \mathcal{W}^{d_0}_{R\rightarrow L}(\hat{\textbf{I}}^R), ..., \mathcal{W}^{d_{D-1}}_{R\rightarrow L}(\hat{\textbf{I}}^R)\}. 
\end{multline}
The TPSV helps the network to distinguish the dynamically-occluded parts in the 3D scene.
Then, we acquire the 3D mask volume by $\textbf{V} = \mathcal{G}_\theta(\Tilde{\textbf{P}})$.

Finally, we can calculate the final MPI $\textbf{M}_o$ by:
\small
\begin{equation}
    \textbf{M}_{o}(x, y, d) = \textbf{M}(x, y, d) \textbf{V}(x, y, d) + \hat{\textbf{M}}(x, y, d) (1-\textbf{V}(x, y, d) ),
\end{equation}
\normalsize
for all $(x, y, d)$.
We define a shorthand version as
\begin{equation}
    \textbf{M}_{o} = \textbf{M} \odot \textbf{V} + \hat{\textbf{M}} \odot (1-\textbf{V}),
\end{equation}
where $\odot$ means element-wise multiplication.
$\textbf{M}_o$ achieves Eq. \ref{eq:remove_disocc} as our learnable mask volume $\textbf{V}$ satisfies Eq. \ref{eq:mask_volume} and we can then render the output image $\textbf{I}_o$ using planar homography and the over composite operation described in Sec. \ref{sec:mpi}. Please refer to Fig.\ref{fig:algorithm} for illustrations.

One major difference between using a 3D mask volume $\textbf{V}(x, y, d)$ and a 2D mask $\textbf{V}'(x, y)$ is that the former is able to segment out the dynamic objects in the 3D space, namely Eq. \ref{eq:mask_volume} and subsequently do Eq. \ref{eq:remove_disocc}.
In Fig. \ref{fig:algorithm}, notice that the mask volume only contains the dynamic object (jumping person in this case).
In contrast, a 2D mask $\textbf{V}'(x, y)$ does not vary with respect to the disparity $d$, making it impossible to manipulate the areas behind dynamic objects.



\edit{
\subsection{Network Architecture}
Our view synthesis pipeline utilizes two different 3D CNNs to predict the MPI volumes and the 3D mask volume as described in Sec.~\ref{sec:3d_mask_volume}.
Both networks have similar structures as the one in Mildenhall et al.\cite{mildenhall2019llff}.
However, we made some adjustments to keep the network light for faster training and less memory consumption.
We show detailed layers for the mask network in Table~\ref{tab:cnn}.
The MPI network has the same structure except for some changes in the overall input and output channels to account for different view counts.

\newcommand\setrow[1]{\gdef\rowmac{#1}#1\ignorespaces}
\newcommand\clearrow{\global\let\rowmac\relax}
\clearrow

\begin{table}[t]
  \centering
  \resizebox{0.95\textwidth}{!}{%
  \begin{tabular}{cccccccc}
    \hline
    \setrow{\bfseries}Layer  & kernel size & stride & dilation & in & out & activation & input \\
    \hline
        conv1\_1  &           7 &      1 &        1 & 12 &   8 &       ReLU &       PSVs \\
        conv1\_2  &           7 &      2 &        1 &  8 &  16 &       ReLU &   conv1\_1 \\
        conv2\_1  &           3 &      1 &        1 & 16 &  16 &       ReLU &   conv1\_2 \\
        conv2\_2  &           3 &      2 &        1 & 16 &  32 &       ReLU &   conv2\_1 \\
        conv3\_1  &           3 &      1 &        1 & 32 &  32 &       ReLU &   conv2\_2 \\
        conv3\_2  &           3 &      2 &        1 & 32 &  64 &       ReLU &   conv3\_1 \\
        conv4\_1  &           3 &      1 &        1 & 64 &  64 &       ReLU &   conv3\_2 \\
        conv4\_2  &           3 &      1 &        1 & 64 &  64 &       ReLU &   conv4\_1 \\
    \hline
             up5  &             &      2 &          &128 & 128 &            &   conv3\_2 + conv4\_2 \\
        conv5\_1  &           3 &      1 &        1 &128 &  32 &       ReLU &      nnup5 \\
        conv5\_2  &           3 &      1 &        1 & 32 &  32 &       ReLU &   conv5\_1 \\
             up6  &             &      2 &          & 64 &  64 &            &   conv2\_2 + conv5\_2 \\
        conv6\_1  &           3 &      1 &        1 & 64 &  16 &       ReLU &      nnup6 \\
        conv6\_2  &           3 &      1 &        1 & 16 &  16 &       ReLU &   conv6\_1 \\
             up7  &             &      2 &          & 32 &  32 &            &   conv1\_1 + conv6\_2 \\
        conv7\_1  &           3 &      1 &        1 & 32 &   16 &      ReLU &      nnup7 \\
        conv7\_2  &           3 &      1 &        1 & 16 &   8 &       ReLU &   conv7\_1 \\
        conv7\_3  &           3 &      1 &        1 &  8 &   1 &    Sigmoid &   conv7\_2 \\
    \hline
  \end{tabular}
  }
  \caption{Details of each layer in our 3D mask network.}
  \label{tab:cnn}
\end{table}
}

\subsection{Loss function}\label{sec:loss_function}
We implement our loss function as a rendering loss, similar to previous work on MPIs \cite{zhou2018stereo, SrinivasanCVPR2019, mildenhall2019llff}.
For the rendering loss, we use view synthesis as the supervision task and let the algorithm render a held-out view from the final MPI $\textbf{M}_o$ (see Fig.\ref{fig:algorithm}).
The rendering loss is as follows:
\begin{equation}
    \mathcal{L} = \frac{||\mathcal{F}_{VGG}(\textbf{I}_o) - \mathcal{F}_{VGG}(\textbf{I}_{gt})||_1}{N},
\end{equation}
where $\mathcal{F}_{VGG}$ is the VGG-19 network \cite{ simonyan2014very}, $N$ is the number of elements in the image $\textbf{I}_o$, and $\textbf{I}_{gt}$ is the held-out ground truth view.
This perceptual loss is similar to the implementation of Chen et al.  \cite{chen2017photographic}.
We also considered a mask supervision loss $\mathcal{L}_m$ and a mask sparsity constraint $\mathcal{L}_s$.
However, we did not find them to be useful for temporal consistency.
Ablation studies on these two losses can be found later in Table \ref{tab:loss}, and details are in Sec.~\ref{sec:ablation}.




\section{Results}\label{sec:experiments}

\begin{figure*}
\begin{center}
\includegraphics[width=1.0\textwidth]{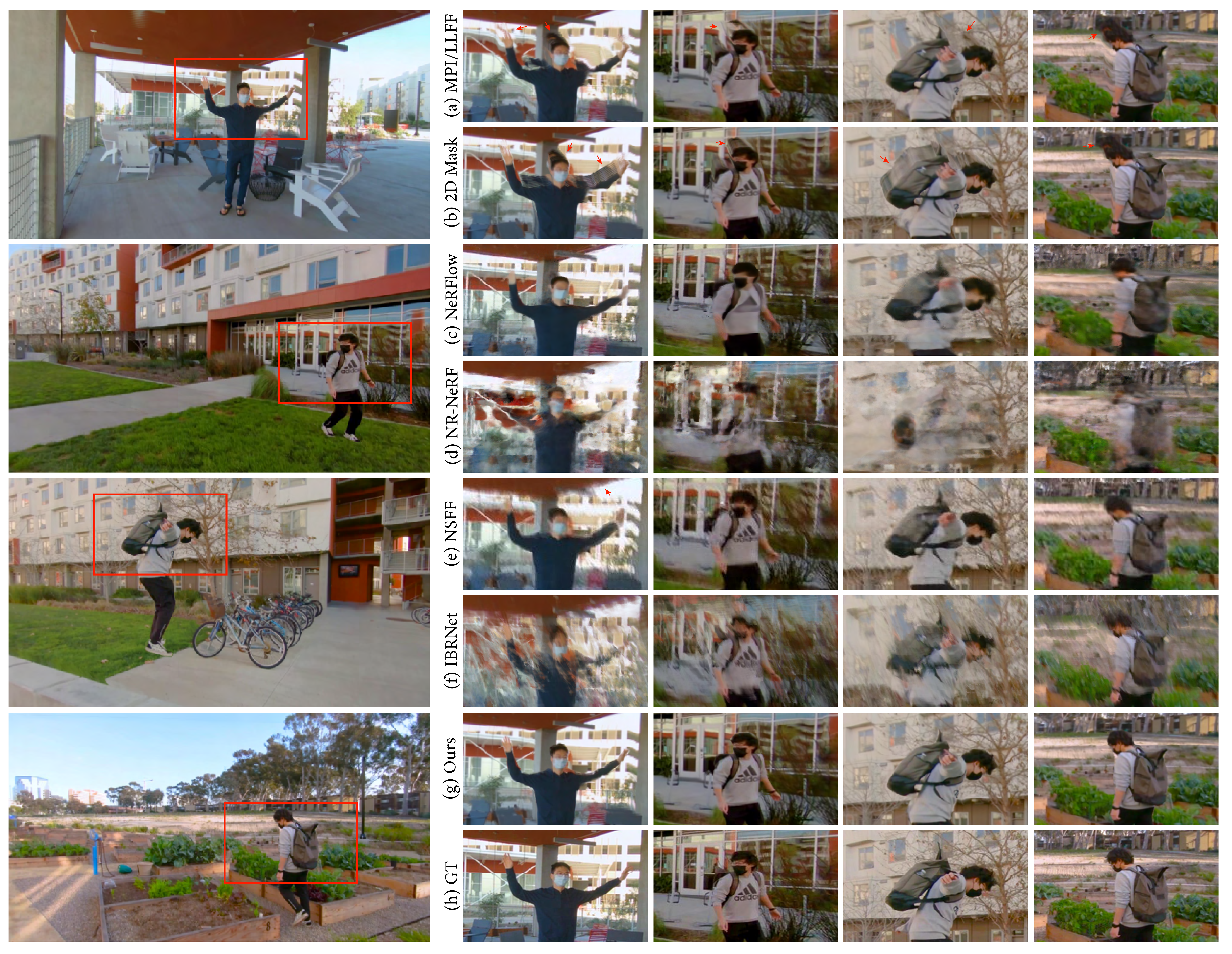}
\end{center}
   \caption{We show visual results on 4 different scenes. These scenes include both fast and slow movements, such as waving, jumping and walking. The novel viewpoint is an extrapolation from the input camera views. \edit{In the above images, each row is rendered using one method from (a) MPI/LLFF~\cite{mildenhall2019llff}, (b) 2D Mask, (c) NeRFlow~\cite{du2021nerflow}, (d) Non-rigid NeRF~\cite{tretschk2021nonrigid}, (e) Neural Scene Flow Field~\cite{li2020neural}, (f) IBRNet~\cite{wang2021ibrnet}, and (g) ours. Last row (h) is the ground truth.} Our proposed method produces results with fewer artifacts and more temporal stability.}
\label{fig:qualitative}
\end{figure*}

\begin{figure*}[t]
\centering
\includegraphics[width=1.0\textwidth]{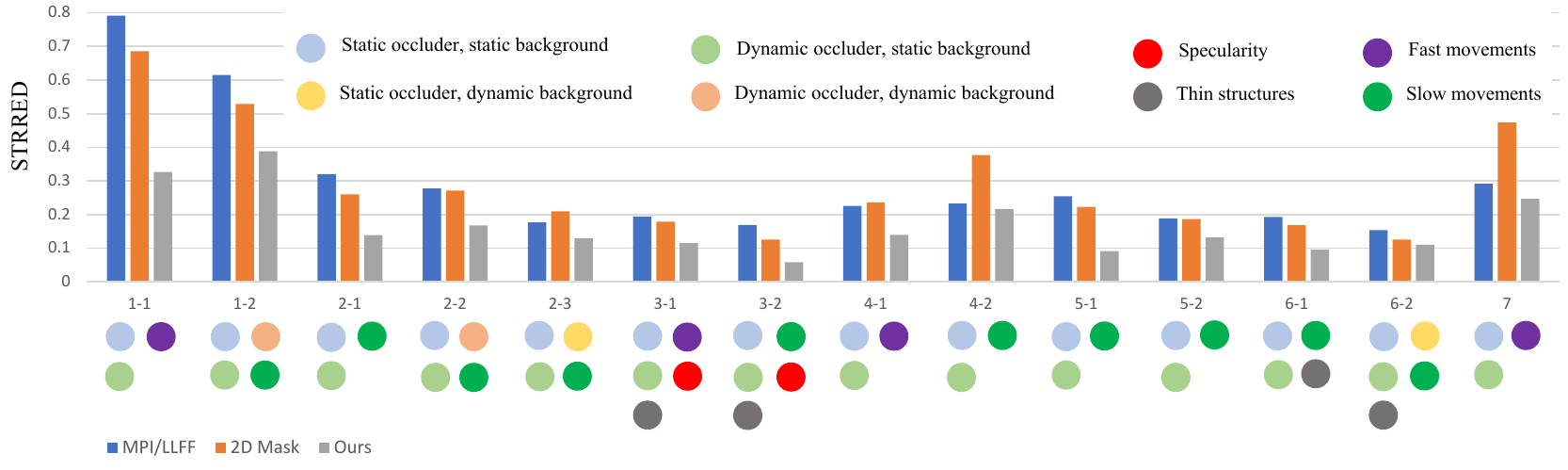}
\caption{STRRED comparison on our dataset with baseline methods. We select 14 clips from 7 different scenes. 1-1, 1-2 denotes clip 1 and clip 2 from scene 1. {\LX{can reduce the empty space at the top of the plot (move the legend lower}}}
\label{fig:analysis}
\end{figure*}

\begin{figure*}
\begin{center}
\includegraphics[width=1.0\textwidth]{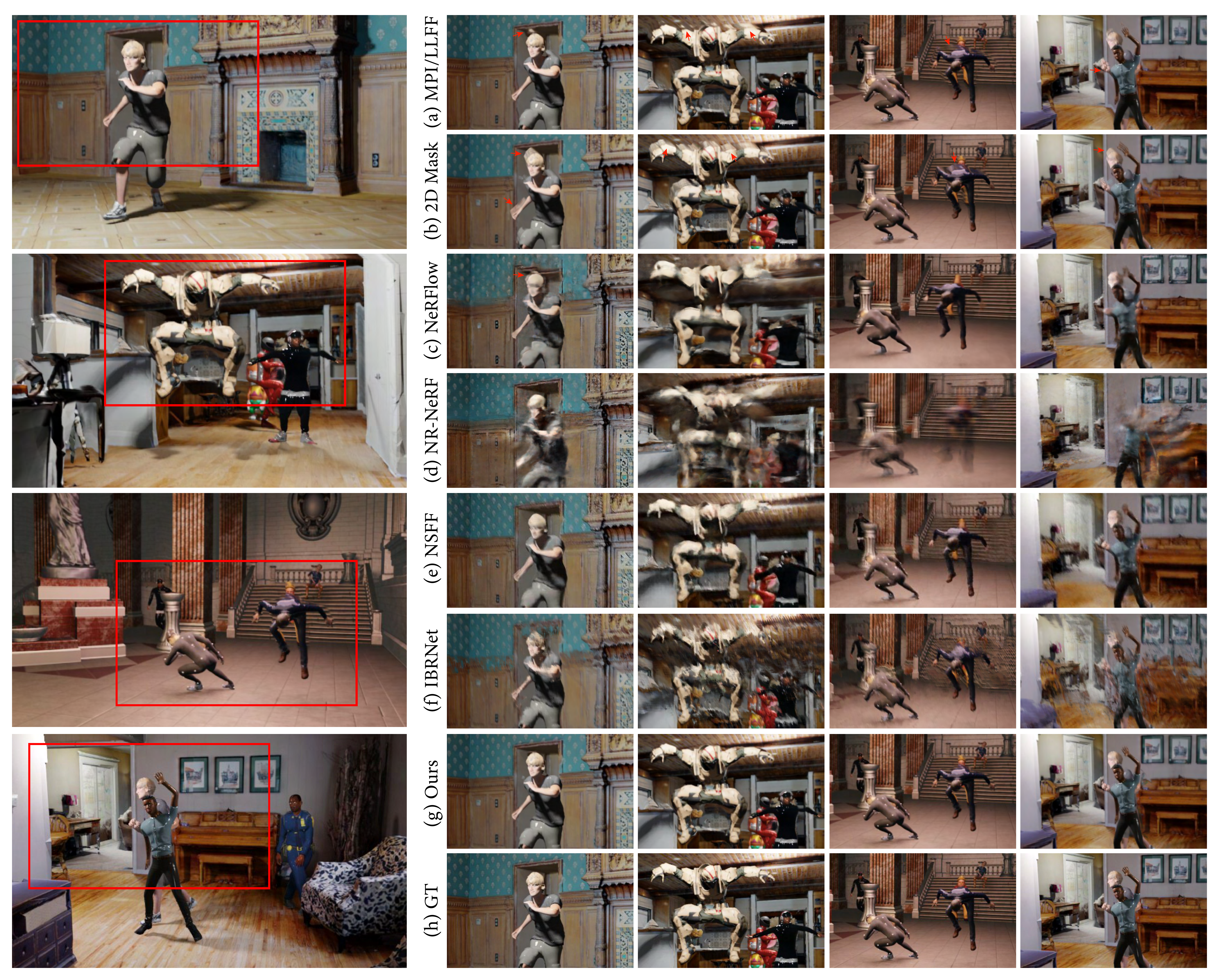}
\end{center}
   \caption{\edit{We show visual results on 4 different synthetic scenes. These scenes include moving characters and dynamic backgrounds. In the above image, each row is rendered using one method from (a) MPI/LLFF~\cite{mildenhall2019llff}, (b) 2D Mask, (c) NeRFlow~\cite{du2021nerflow}, (d) Non-rigid NeRF~\cite{tretschk2021nonrigid}, (e) Neural Scene Flow Field~\cite{li2020neural}, (f) IBRNet~\cite{wang2021ibrnet}, and (g) ours. Last row (h) is the ground truth. Our proposed method (g) produces results with fewer artifacts and more temporal stability.}}
\label{fig:qualitative-syn}
\end{figure*}

In this section, we discuss implementation details for our network in Sec. \ref{sec:implementation}.
Then we show comparisons to other methods on our dataset in Sec. \ref{sec:comparison}.
\edit{
We include comparisons on our synthetic dataset in Sec.~\ref{sec:synthetic}.
To explore the effects of different loss functions, we show the ablation studies in Sec.~\ref{sec:ablation}.
We show that our method is able to degrade gracefully even doing view extrapolation far outside the viewing volume in Sec.~\ref{sec:large_dist}.
Our method can also be extended to incorporate more input views in Sec.~\ref{sec:more_views}.
}
Finally we discuss limitations of our current setup and method in Sec. \ref{sec:limitations}.
Result videos can be found in the supplementary materials.

\subsection{Implementation details}\label{sec:implementation}
Due to GPU memory constraints, we choose a two-step training scheme to train our network.
We first train the MPI network on the RealEstate10K dataset \cite{zhou2018stereo}, and then train only the mask network on our own video dataset.
This training scheme can keep the memory usage within a reasonable range and the speed fast enough.

The MPI generation network is trained by predicting a held-out novel view and applying the rendering loss $\mathcal{L}$ as supervision.
This stage is trained for 800K steps.
After the previous pretraining stage, we freeze the weights of the MPI network and train only the mask network using the loss $\mathcal{L}$.
The network takes 2 random views from the 10 views as input and we randomly choose a target camera position from the rest of the views at each step.
We select 86 out of the 96 scenes as our training dataset and images are rescaled to 640$\times$360.
This second stage is trained for 100K steps.
The learning rate is set to $5e-4$ for both stages.
Our training pipeline is implemented in PyTorch\cite{NEURIPS2019_9015} and training takes around 5 days on a single RTX 2080Ti GPU.
With resolution in 640$\times$360, inferencing $\textbf{M}_o$ using our full pipeline takes around 1.75s, while rendering takes another 0.28s.
Note that the rendering pipeline is implemented in PyTorch without further optimization.
In practice, it could be significantly faster with OpenGL or other rasterizer.

\subsection{\edit{Comparisons on real data}}\label{sec:comparison}
For comparison, we choose 7 unseen videos from the dataset and subdivide them into 14 clips, focusing on salient movements in the scene. \edit{The methods we chose to evaluate includes MPI-based methods like LLFF, and also emerging NeRF-based methods like Nonrigid-NeRF, Neural Scene Flow Field, and NeRFlow}
We ran all methods on the clips with camera 4 and 5 as input and others as the target output (see Fig. \ref{fig:camera_rig}).
Error metrics are calculated between the output and the ground truth images.
\edit{For monocular NeRF-based methods~\cite{tretschk2021nonrigid,li2020neural,du2021nerflow}, as they assume the input to be monocular, moving camera, and have increasing time steps, we alternate between left and right views to satisfy this assumption.
This allows the algorithm to treat the input as a monocular video with the camera jumping between two viewpoints.
}

We compare with 6 baseline approaches:
(1) MPI/LLFF is our adaptation of Mildenhall et al.\cite{mildenhall2019llff} to work with only two input views and different camera intrinsics.
It processes the stereo input videos and renders the novel view frames on a per-frame basis.
(2) 2D mask is our naive baseline method, which is similar to our pipeline, except that it uses a foreground mask $\textbf{V}'(x, y)$ generated by the background matting method\cite{BGMv2} with $\textbf{I}$ and $\hat{\textbf{I}}$ as inputs.{\LX{Should clarify that this 2D mask method is our own custom baseline method, rather than an existing approach.}}
The blended MPI $\textbf{M}_o'$ for (2) is obtained by
\begin{equation*}
    \textbf{M}_o' =  \textbf{V}'(x, y) \odot \textbf{M} +
        (1-\textbf{V}'(x, y)) \odot \hat{\textbf{M}},
\end{equation*}
where the 2D mask has been expanded into 3D by repeating its values along the depth dimension.
(3) IBRNet~\cite{wang2021ibrnet} uses the official implementation and takes 2 views as input on a per-frame basis.
\edit{
(4) NeRFlow~\cite{du2021nerflow} uses the official implementation and we slightly modify the necessary parts to allow for two alternating views as input.
(5) NSFF~\cite{li2020neural} is also adapted from the official implementation to take two input views.
(6) Non-rigid NeRF~\cite{tretschk2021nonrigid} uses the released official implementation with modifications to enable two-view inputs.
For (4)-(6), we train them for 20,000 steps for each scene and render the corresponding viewpoints.
}
Please refer to our supplementary materials for the video results.

\begin{table}[t]
 {\caption{Comparison on our evaluation dataset. We compare with different baseline methods and the results show that our 3D mask offers much better temporal stability. 2D mask does not improve much because it fails to resolve the ambiguity in disoccluded areas.}
  \label{tab:comparison}}
  \resizebox{0.95\columnwidth}{!}
  {\begin{tabular}{c | c c c c}
    \textbf{Methods} & \textbf{Mask} & \textbf{STRRED}$\downarrow$ & \textbf{PSNR}$\uparrow$ & \textbf{SSIM}$\uparrow$\\
    \hline
    MPI/LLFF \cite{mildenhall2019llff} & No Mask & 0.2917 & 25.52 & 0.8227  \\
    2D Mask & 2D & 0.2892 & 25.50 & 0.8242 \\
    IBRNet (2-view) \cite{wang2021ibrnet} & No Mask & 2.2606 & 21.49 & 0.6713 \\
    \edit{NeRFlow} \cite{du2021nerflow} & \edit{No Mask} & \edit{3.2646} & \edit{16.8081} & \edit{0.4146} \\
    \edit{NSFF} \cite{li2020neural} & \edit{No Mask} & \edit{1.4230} & \edit{17.0368} & \edit{0.4197} \\
    \edit{Non-rigid NeRF} \cite{tretschk2021nonrigid} & \edit{No Mask} & \edit{2.3941} & \edit{18.1070} & \edit{0.4997} \\
    \textbf{Ours} & 3D & \textbf{0.1683} & \textbf{26.22} & \textbf{0.8390}  \\

    \hline
  \end{tabular}}
 
\end{table}

\begin{figure*}
\begin{center}
\includegraphics[width=0.95\textwidth]{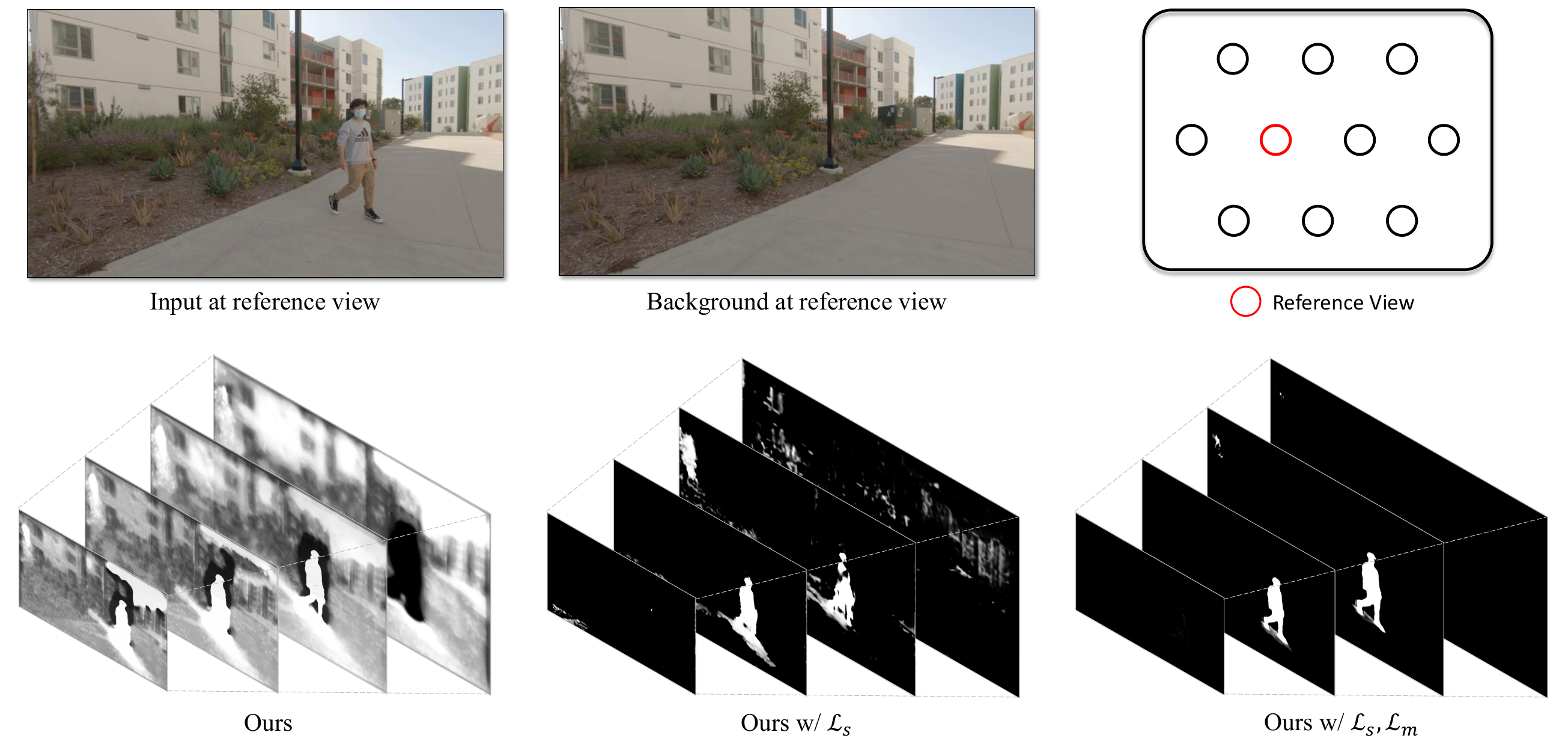}
\end{center}
   \caption{\edit{3D visualization of the masks from different loss functions. With alpha values from the instantaneous MPI, we collapse the mask volumes using over composite to reduce plane count from 32 to 4 for better visualization. (e.g. plane 1$\sim$8 to the furthest plane, ..., plane 25$\sim$32 to the nearest plane.)
   Note that there is no supervision on static parts in our final loss function, so the values in those parts are unconstrained, resulting in soft blending between instantaneous frames and the background.
   In general, the 3D mask achieves better temporal consistency by replacing the erroneous disoccluded parts with correct background observations.
   }
   }
\label{fig:mask3d}
\end{figure*}

From Table \ref{tab:comparison}, we see that our method is able to achieve temporally-coherent rendering, while offering better visual quality and fewer distortions.
Specifically, we employ the STRRED metric \cite{soundararajan2012video} to evaluate stability across time.
Our method significantly reduces the temporal artifacts across most scenes while also keeping PSNR and SSIM better than the baseline methods.
For MPI/LLFF, since it does not utilize the information across the whole video, it yields more flickering and distorted areas as can be seen in Fig. \ref{fig:qualitative}.
For example, in the top scene, there is a ghosting artifact around the person's head and it changes frame-by-frame, resulting in flickering video.
The 2D mask method is a binary mask that naively selects the dynamic parts in $\textbf{M}$ and the background in $\hat{\textbf{M}}$ to produce the final MPI.
As a result, it amplifies the stack of cards artifacts (see Fig. \ref{fig:qualitative}) and also slightly worsens the visual quality as shown in Table \ref{tab:comparison}.
IBRNet~\cite{wang2021ibrnet}, does not work well with 2-view input and it produces poor results compared to ours.
\edit{Concurrent monocular NeRF-based methods~\cite{tretschk2021nonrigid,du2021nerflow,li2020neural} perform similarly in Table~\ref{tab:comparison}. With only two input viewpoints, they fail to represent even the static scene components since there are not enough multi-view cues for reconstruction.
For dynamic parts of the scene, NSFF provides more stable quality as can be seen from the STRRED metric.
In general, our proposed method provides state-of-the-art performance over other previous and concurrent work.}
\edit{
We show qualitative results in Fig. ~\ref{fig:qualitative}.
Each inset column corresponds to a scene as shown on the leftmost side.
We show the MPI baseline method in row (a) and 2D mask baseline in row (b).
These two methods suffer from stack-of-card artifacts in particular in the disoccluded regions.
2D mask fails to solve the problem and sometimes makes it more apparent.
This is because 2D mask does not reason about the 3D geometry of the scene.
For the more recent NeRF-based methods, we show them in row (c-f).
NeRFlow~\cite{du2021nerflow} provides better static scene reconstruction than other methods.
However, it produces blurred results and lacks high-frequency details as can be seen from the second image in row (c).
On the other hand, our proposed method is able to make the text on the person more legible and sharper, while suffering little to no disocclusion artifacts.
Non-rigid NeRF~\cite{tretschk2021nonrigid} suffers from significant artifacts when rendering the images.
This is possibly due to sparse viewpoints and the network is trying to compensate with deformation fields.
NSFF~\cite{li2020neural} generates sharper images than NeRFlow, but it suffers from blurriness in static parts of the scene.
IBRNet~\cite{wang2021ibrnet} produces noisy results given two input views on a frame-by-frame basis.
Their method tries to blend different viewpoints with a ray transformer to synthesize disoccluded regions.
However, given two input views, this becomes even more difficult because of the lack of samples.}

To further analyze how temporal consistency is affected, we characterize the clips with different properties including different types of occlusion discussed in Sec. \ref{sec:statistics} and show the results in Fig. \ref{fig:analysis}.
As stated earlier, several clips are selected from the 7 scenes to show salient motions.
\edit{
We only include results from MPI/LLFF~\cite{mildenhall2019llff}, 2D mask and Ours, as other methods have significantly higher STRRED.
}
From the results, we observe that faster movements could often result in worse temporal consistency, like the differences between clip 1-1 and 1-2.
There is an interesting failure in 4-2 for the 2D mask method.
4-1 is the jumping scene in Fig. \ref{fig:qualitative}, and 4-2 shows a person walking in the same scene.
Although the movement is slower, the person walks past several areas with large appearance changes in 4-2.
As a result, the artifacts in the 2D mask are much more obvious, and the video flickers more than other methods, leading to a worse STRRED score.




\subsection{\edit{Comparisons on synthetic data}}\label{sec:synthetic}
\edit{In addition to real data, we also crafted a synthetic dataset and tested different methods on it. The synthetic dataset not only can provide us real ground truth to make proper comparisons, but also can illustrate scenes and movements hard to capture in real life, for example, complex moving backgrounds. The synthetic dataset is constructed using scenes from the Habitat-Matterport 3D dataset~\cite{ramakrishnan2021hm3d} and UE4 Sun Temple~\cite{OrcaUE4SunTemple}, and the moving characters in the scene are pre-animated characters from Adobe Mixamo.
We used Blender~\cite{blender2020} to composite the scenes, and replicated the 10-view camera array with parameters similar to our GoPro setup.
We deliberately set all the cameras to have the same camera intrinsics in order to reduce unwanted artifacts.

For each scene, we rendered 60 frames of the animation, and produced camera poses for all 10 cameras.
As all the camera poses can be directly obtained from Blender, we do not need COLMAP~\cite{schoenberger2016mvs, schoenberger2016sfm} to estimate camera poses anymore.
The background images are still obtained using median filter. Similar to our evaluation on the real dataset, we chose cameras 4 and 5 as input. In Table~\ref{tab:synthetic}, we show the numbers of various methods.
The proposed method achieves favorable results compared to other baselines.
Additionally, we show qualitative results in Fig. ~\ref{fig:qualitative-syn}.
For MPI/LLFF, the numbers are slightly worse than our proposed algorithm, because the main difference is in the disoccluded regions.
It can be seen in the row (a) around the moving characters.
2D mask introduces more artifacts and thus results in worse numbers across all metrics.
In row (b), 2D mask exacerbates the artifacts and creates more visible repeated texture in the disoccluded regions.
NeRF-based methods perform slightly better on the synthetic dataset, as the camera parameters are more precise.
However, they still fail to produce sharp imagery.
For example, NeRFlow lacks the details on the leftmost character in the third column in row (c).
Furthermore, the second column in row (d) shows blurriness and ghosting artifacts for Non-rigid NeRF.
NSFF (e) has issues rendering complex static scene texture in the last column.
The table to the left shows distorted edges compared to our proposed method.
IBRNet (f) still generates renderings with heavy distortions, even though the coarse geometry seemingly matches the ground truth.
Our method (g) provides the best visual result and it is able to generalize to unseen synthetic scenes when trained on real data.
Please refer to the supplementary video for more results.
}

\begin{table}[t]
 {\caption{\edit{Comparison on the synthetic evaluation dataset.}}
  \label{tab:synthetic}}
  \resizebox{0.95\columnwidth}{!}
  {\begin{tabular}{c | c c c c}
    \textbf{Methods} & \textbf{Mask} & \textbf{STRRED}$\downarrow$ & \textbf{PSNR}$\uparrow$ & \textbf{SSIM}$\uparrow$\\
    \hline
    \edit{MPI/LLFF} \cite{mildenhall2019llff} & \edit{No Mask} & \edit{0.2889} & \edit{26.1167} & \edit{0.8345}  \\
     \edit{2D Mask} & \edit{2D} & \edit{0.5428} & \edit{24.0146} & \edit{0.8082}  \\
     \edit{IBRNet (2-view)} \cite{wang2021ibrnet} & \edit{No Mask} & \edit{1.7984} & \edit{21.3727} & \edit{0.6942}  \\
    \edit{NeRFlow} \cite{du2021nerflow} & \edit{No Mask} & \edit{1.8306} & \edit{19.9902} & \edit{0.5996} \\
    \edit{NSFF} \cite{li2020neural} & \edit{No Mask} & \edit{1.0627} & \edit{19.7176} & \edit{0.5577} \\
    \edit{Non-rigid NeRF} \cite{tretschk2021nonrigid} & \edit{No Mask} & \edit{3.1401} & \edit{18.9230} & \edit{0.5947} \\
    \edit{\textbf{Ours}} & \edit{3D} & \edit{\textbf{0.2812}} & \edit{\textbf{26.1348}} & \edit{\textbf{0.8342}}  \\

    \hline
  \end{tabular}}
 
\end{table}

\subsection{\edit{Ablation Studies on Loss Function}}\label{sec:ablation}
\edit{

In this sub-section, we experiment with different losses to see if we can acquire a 3D mask volume that is more interpretable and possesses physical meaning.
Two additional loss functions are described as follows.
The first loss is a mask supervision loss $\mathcal{L}_m$, which forces the mask volume to match the shape of the dynamic object in the scene.
The second loss is a sparsity loss $\mathcal{L}_s$ applied on the mask volume to encourage the network to reuse $\hat{\textbf{M}}$ more.
To be more specific, for the mask loss, we use the work by Lin et al.\cite{BGMv2}, which takes the individual frame $\textbf{I}$ and the background $\hat{\textbf{I}}$ in the video to generate a dynamic object mask $\textbf{V}_{gt}$ we later use as supervision. 
To supervise the mask volume, we directly regularize the over-composited alphas from the warped foreground MPI volume $\mathcal{W}(\textbf{M}\odot \textbf{V})$ to be consistent with $\textbf{V}_{gt}$.
We denote the over-composited alpha values as $m_1$. 
This mask loss is similar to the mask supervision loss in Lu et al. \cite{10.1145/3414685.3417760}
We calculate the estimated background mask $m_0$ by dilating the foreground mask with a kernel of size $(5, 5)$ to produce $m_1'$.
The background mask is then $m_0 = 1-m_1'$.
And the mask supervision loss is:
\begin{equation}
    \mathcal{L}_{m} = \frac{||m_1\odot (1-\textbf{V}_{gt})||_1}{2||m_1||_1} + \frac{||m_0\odot \textbf{V}_{gt}||_1}{2||m_0||_1}.
\end{equation}

Another loss is a $L_1$ sparsity constraint on the mask volume to ensure it only covers the necessary portions,
\begin{equation}
    \mathcal{L}_{s} = ||\sum_{(x, y, d)}{\textbf{V}(x, y, d)}||_1.
\end{equation}
We use $\mathcal{L} + 0.1 \mathcal{L}_s + 0.25 \mathcal{L}_m$ for the full combination and $\mathcal{L} + 0.1 \mathcal{L}_s$  for the additional sparsity constraint.

}

\begin{table}[t]
  {\caption{Effect of different loss functions. Our rendering loss offers better temporal consistency and slightly better visual quality.  }\label{tab:loss}}
  {\begin{tabular}{c | c c c c c}
    \textbf{Methods} & \textbf{STRRED}$\downarrow$ & \textbf{PSNR}$\uparrow$ & \textbf{SSIM}$\uparrow$ \\
    \hline
    Ours &  \textbf{0.1683} & \textbf{26.22} & 0.8390 \\
    Ours w/ $\mathcal{L}_s$ & 0.1745 & 26.18 & \textbf{0.8393} \\
    Ours w/ $\mathcal{L}_s, \mathcal{L}_m$ & 0.1900 & 26.09 & 0.8374 \\
    \hline
  \end{tabular}}
\end{table}

\begin{figure*}
\begin{center}
\includegraphics[width=0.8\textwidth]{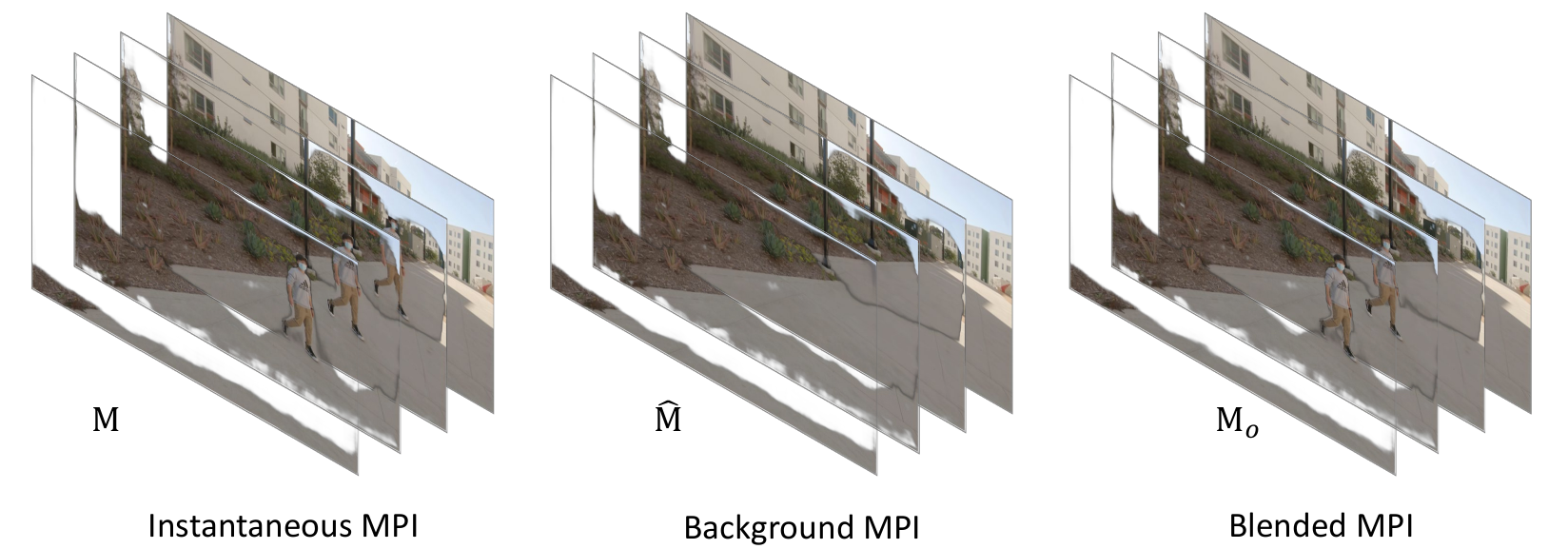}
\end{center}
   \caption{\edit{3D visualization of the MPI volumes using our loss function $\mathcal{L}$. Note that the person on the furthest plane in $\textbf{M}$ is replaced by the background in $\textbf{M}_o$.}}
\label{fig:3dmpi}
\end{figure*}

\edit{
As shown in Table \ref{tab:loss}, our rendering loss still offers the most temporally-stable results, whereas the other two losses  trade temporal consistency for better interpretability.
It is reasonable that the mask supervision loss helps the network to give a sparser and tighter prediction on the dynamic objects.
However, it does not take into account the movements of the foliage and the shadows, producing slightly unstable results in those areas.
The sparsity constraint is able to achieve marginally better quality than the full $\mathcal{L}_s, \mathcal{L}_m$ combination as it retains some parts of the scene which might cover the slight differences between frames.

Mask visualization can be found in Fig. ~\ref{fig:mask3d}.
From the figure, we can observe that our mask volume removes areas around the edges of the dynamic object and the occluded areas behind it.
Moreover, the mask softly blends the shadows cast by the moving object.
Adding $\mathcal{L}_s$, the mask becomes sparser, ignoring most static areas.
However, as shown in Fig. ~\ref{fig:mask3d}, it still contains some areas around the plants on the left and the building in the back.
With $\mathcal{L}_s, \mathcal{L}_m$, the mask has more physical meaning and the resulting 3D mask only covers the dynamic object.
This might be useful to extract moving objects for other uses such as editing or object insertion.

We further examine the 3D visualization of $\textbf{M}, \hat{\textbf{M}}, \text {and } \textbf{M}_o$ in Fig. ~\ref{fig:3dmpi}.
Note that in the blended MPI $\mathbf{M}_o$, the occluded area behind the person is filled with actual background information, unlike in $\textbf{M}$, which has repeated texture of the dynamic object. 
Since we do not enforce any constraints on the static parts of the scene, our mask has random values in these areas and softly blends them with the background MPI.
\textit{This does not affect temporal consistency too much as the difference is minor and some areas are free space which does not contribute any color to the MPI volume as shown in Fig. ~\ref{fig:3dmpi}.}
}

\begin{figure*}
\begin{center}
\includegraphics[width=0.95\textwidth]{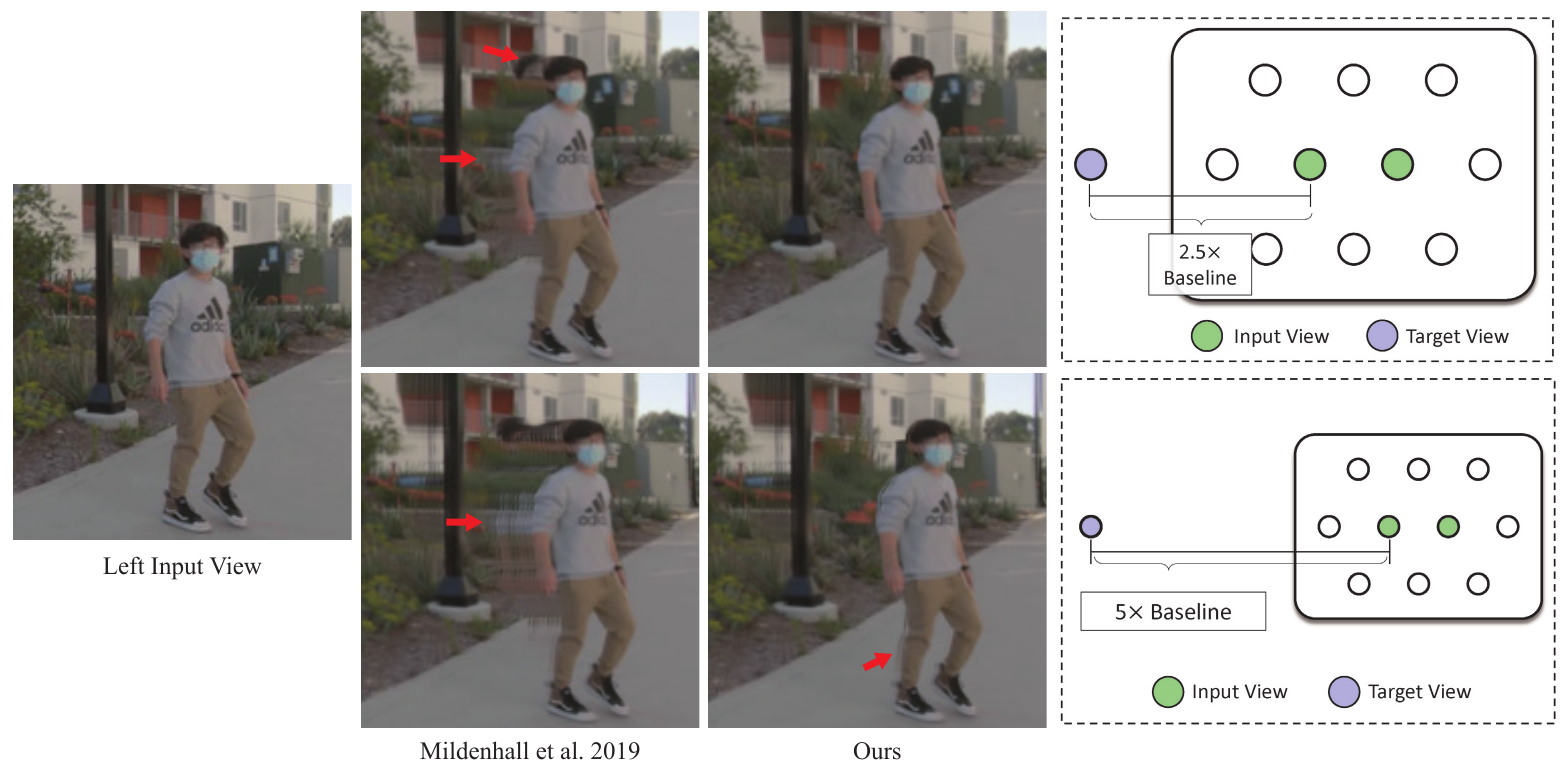}
\end{center}
   \caption{\edit{Our algorithm is able to provide better visual quality than baseline methods even when the novel viewpoint is far away from the input view. We show results when the baseline is $2.5\times$ and $5\times$ baseline between input views. Note that in the $5\times$ case, our method produces fewer artifacts compared to Mildenhall et al.\cite{mildenhall2019llff}, offering a more graceful degradation.}}
\label{fig:extrapolation}
\end{figure*}

\begin{figure*}
\begin{center}
\includegraphics[width=\textwidth]{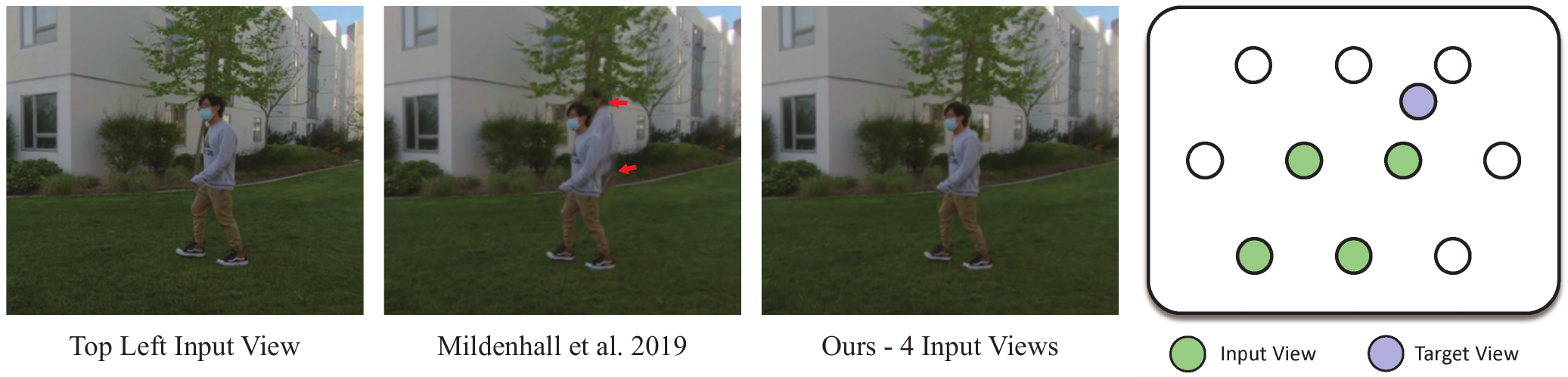}
\end{center}
   \caption{\edit{Our proposed method can also be extended to take 4-view input. We feed 4 input views to both the MPI and mask networks to acquire our result. Here the baseline method is also adjusted to use 4 input views instead of 2. Notice that the artifacts around the person do not appear in our result.}}
\label{fig:4views}
\end{figure*}

\subsection{\edit{Large distance view extrapolation}}\label{sec:large_dist}
\edit{
In Fig. ~\ref{fig:extrapolation}, we show results when the target camera is translated far more than the baseline of the input camera pair.
When large translational movement is introduced, the conventional method\cite{mildenhall2019llff} starts to show artifacts in the disoccluded regions.
On the contrary, our method still preserves the background details even when the motion is larger, offering a more graceful reduction in quality as the distance is increased.
}
\subsection{\edit{Extension to more input views}}\label{sec:more_views}
\edit{
Although our proposed method primarily targets binocular view extrapolation, we also demonstrate that it can be extended to utilize more input views in Fig. ~\ref{fig:4views} and in the supplementary video.
With more input views, it can acquire better scene geometry for some cases where there are ambiguities in the plane sweep volume.
For example, some ambiguities might occur when there is straight texture-less structure (beams or handrails) parallel to the camera baseline.
Using additional cameras can provide more geometric information and avoid similar situations.
In Fig. ~\ref{fig:4views}, the main difference is that we modify our network to take 4 input views, which convert to 4 instantaneous images and 4 background images as input to the mask network, and output the 3D mask volume as in the pipeline shown in Fig. ~4 in the main paper.
}

\subsection{Limitations}\label{sec:limitations}
\begin{figure}
\begin{center}
\includegraphics[width=1.0\textwidth]{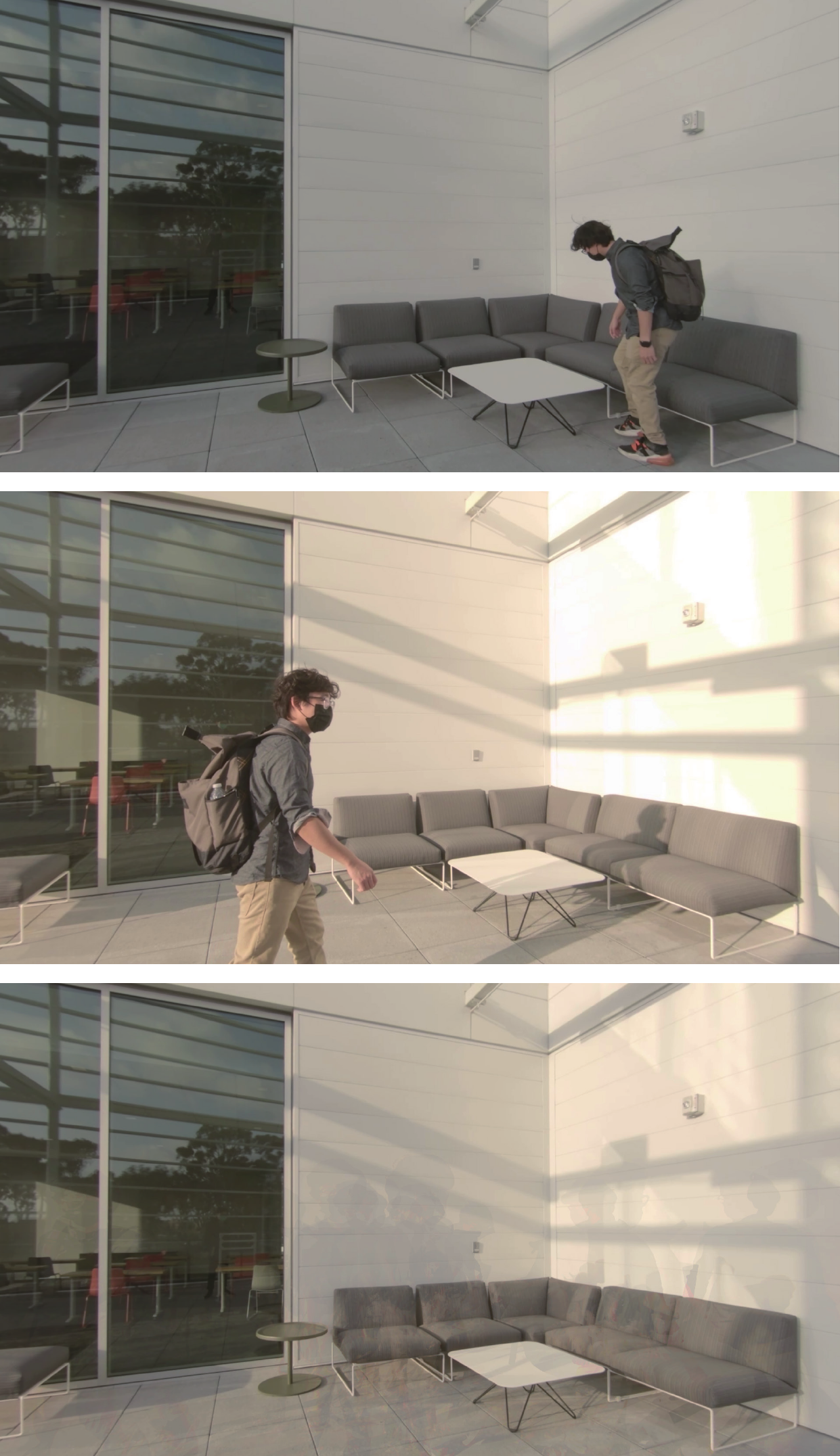}
\end{center}
\vspace{-1em}
   \caption{\edit{From top to bottom, we show frame 0, frame 966 (last frame) and the extracted background. Since the lighting changes drastically during this scene, the extracted background contains a lot of ghosting artifacts.}}
\label{fig:background}
\end{figure}

The proposed dataset and algorithm have a few limitations:
First, we limit our camera to stay static when capturing.
This is mainly due to the limitations of synchronization and pose estimation.
Although we can achieve good synchronization with software-based methods, there are still a few milliseconds of error.
This error could be magnified when the camera rig is in motion and lead to bad estimates of the camera poses.
The camera poses across time would also require more calculations, possibly leading to accumulating errors in the system.
These issues could be solved by calibrating the camera trajectory of one of the cameras and utilizing the rigid assumption to infer the trajectories of other cameras.
Another limitation is that we require an estimate of the static background.
This is easily achievable by applying a median filter.
While it works for most of the scenes, this method is sometimes not reliable.
We show one example in Fig. ~\ref{fig:background}.
In this particular case, the sun light appears after a while in the video, casting hard shadows on the walls.
As a result, the background is difficult to determine.
Another possible case happens when a static object is moved during the video.
It is ambiguous to define the exact background for this case as both states might take up a large portion of the video.
Thus, it might require more careful division of different states or using a lighting-agnostic method.
There are more advanced approaches\cite{he2012incremental,hauberg2014grassmann} that can be used in the future.

\section{Conclusions and Future Work}

In this paper, we discuss view synthesis of dynamic scenes with stereo input videos.
The main challenge is that rendered results are prone to temporal artifacts like flickering in the disoccluded regions.
To tackle this issue, we introduce a novel 3D mask volume extension to carefully replace the disoccluded areas with background information acquired from the temporal frames.
Additionally, we introduce a high-quality multi-view video dataset, which contains 96 scenes of various human interactions and outdoor environments shot in 120FPS.

In future work, we would like to extend our dataset and method to consider dynamic camera motions, and to operate on even larger baselines.
In summary, we believe video view synthesis for dynamic scenes is the next frontier for immersive applications, and this paper has taken a key step in that direction.

\section{Acknowledgement}

This work was supported in part by a Qualcomm FMA Fellowship, ONR grant N000142012529, ONR grant N000141912293, NSF grant 1730158, all awarded to the UCSD researchers.
We also acknowledge gifts from Adobe, Google, Amazon, a Sony Research Award, an Amazon Research award, a Facebook Distinguished Faculty Award, the Ronald L. Graham Chair, and the UC San Diego Center for Visual Computing.
Part of the work was done when KEL was an intern at Facebook.
Lastly, we thank Jiyang Yu, Yuzhe Qin, Dominique Meyer, Eric Lo, Thomas DeFanti, Jürgen Schulze and Michael Broxton for comments on hardware setup.
\newpage


%





\ifCLASSOPTIONcaptionsoff
  \newpage
\fi



%



\Urlmuskip=0mu plus 1mu
\bibliographystyle{IEEEtran}
\bibliography{egbib}

\end{document}